\documentclass[10pt,twocolumn,letterpaper]{article}

\usepackage[protrusion=true,expansion=true]{microtype}{}

\usepackage[pagenumbers]{cvpr} 

\usepackage{tabularx}
\newcolumntype{C}{>{\centering\arraybackslash}X}

\definecolor{cvprblue}{rgb}{0.21,0.49,0.74}
\usepackage[pagebackref,breaklinks,colorlinks,allcolors=cvprblue]{hyperref}

\usepackage[capitalize]{cleveref}
\crefname{section}{Sec.}{Secs.}
\Crefname{section}{Section}{Sections}
\Crefname{table}{Table}{Tables}
\crefname{table}{Tab.}{Tabs.}

\IfFileExists{microtype.sty}{\usepackage[protrusion=true,expansion=true]{microtype}}{}

\usepackage{pifont} 
\usepackage{multirow}
\usepackage{float}
\usepackage{graphicx}
\usepackage{amsmath}
\usepackage{amssymb}
\usepackage{booktabs}
\usepackage[table]{xcolor} 
\definecolor{solaHL}{HTML}{FFF7CC} 

\usepackage[pagebackref,breaklinks,colorlinks,allcolors=cvprblue]{hyperref}

\usepackage[capitalize]{cleveref}
\crefname{section}{Sec.}{Secs.}
\Crefname{section}{Section}{Sections}
\Crefname{table}{Table}{Tables}
\crefname{table}{Tab.}{Tabs.}
\definecolor{cvprblue}{rgb}{0.21,0.49,0.74}
\def\paperID{5280} 
\def\confName{CVPR}
\def\confYear{2026}

\title{SoLA-Vision: Fine-grained Layer-wise Linear Softmax Hybrid Attention}

\author{
Ruibang Li\textsuperscript{1,2,3},
Guan Luo\textsuperscript{1,2,3}\thanks{Corresponding author.},
Yiwei Zhang\textsuperscript{1,2,3},
Jin Gao\textsuperscript{1,2,3},
Bing Li\textsuperscript{1,2,3},
Weiming Hu\textsuperscript{1,2,3,4}\\
\textsuperscript{1}State Key Laboratory of Multimodal Artificial Intelligence Systems (MAIS), CASIA\\
\textsuperscript{2}School of Artificial Intelligence, University of Chinese Academy of Sciences\\
\textsuperscript{3}Beijing Key Laboratory of Super Intelligent Security of Multi-Modal Information\\
\textsuperscript{4}School of Information Science and Technology, ShanghaiTech University\\
\small\ttfamily{\{ liruibang2024, zhangyiwei2023\}@ia.ac.cn,
\{ gluo, jin.gao, bli, wmhu\}@nlpr.ia.ac.cn}
}

\begin{document}

\maketitle

\begin{abstract}

Standard softmax self-attention excels in vision tasks but suffers from $\mathcal{O}(N^2)$ complexity, limiting high-resolution implementation. Linear attention reduces memory and computational cost to $\mathcal{O}(N)$, yet its compressed states often impair modeling capacity and performance. In this work, we present an {analytical study} comparing the core properties of linear and softmax attention on visual representation learning from a fresh \textbf{layer-stacking perspective}. We further conduct a systematic empirical experiments on the \textbf{layer-wise}{ hybridization patterns} of linear and softmax attention. The results indicate that, in contrast to rigid intra-block hybrid designs, fine-grained layer-wise hybridization achieves comparable or even superior performance while requiring fewer softmax attention layers. Building on this, we propose \textbf{SoLA-Vision (Softmax-Linear Attention Vision)}, a flexible \textbf{layer-wise hybrid attention} backbone with fine grained control over the integration of linear and softmax attention. Through strategically inserting a minimal number of global softmax layers in a layer-wise manner, SoLA-Vision achieves a compelling trade-off between performance and computational cost. For classification on the ImageNet-1K dataset, \textbf{SoLA-Vision} outperforms purely linear and other hybrid attention models. In dense prediction tasks, SoLA-Vision consistently surpasses strong baselines by a considerable margin, due to its flexible layer-wise strategy for global softmax attention hybridization. Code will be released.

\end{abstract}

\section{Introduction}
\label{sec:intro}

For years, {Vision Transformers (ViTs)}~\cite{dosovitskiy2020image} have been a leading paradigm in visual representation learning. Their success is largely attributed to the global contextual modeling of the {attention mechanisms}~\cite{vaswani2017attention}. However, the \textbf{quadratic computational complexity} ($\mathcal{O}(N^2)$) of standard softmax self-attention severely limits their application to high-resolution images, creating a significant computational bottleneck. To mitigate this $\mathcal{O}(N^2)$ cost, two main research lines have emerged. The first, window-based local attention~\cite{liu2021swin}, restricts softmax attention computations to local partitions. While efficient, this constrains the effective receptive field and impedes long-range dependency modeling~\cite{yang2021focal,xu2025mswa}. The second approach employs linear complexity attention~\cite{katharopoulos2020transformers, zhu2024vision}, which achieves $\mathcal{O}(N)$ efficiency by compressing information into a hidden state. However, this cumulative compression often leads to information decay and representational bottlenecks, struggling to preserve fine-grained spatial details (as we analyze in \Cref{sec:prelim}).

\begin{figure}
    \centering
    \includegraphics[width=1\linewidth]{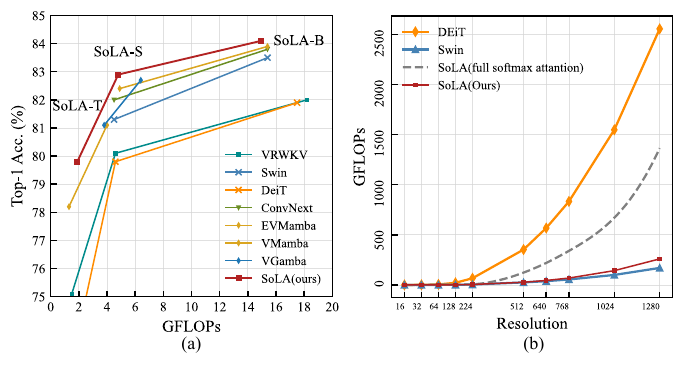}
    \vspace{-20pt}
    \caption{\textbf{Performance and efficiency of SoLA-Vision.} (a) ImageNet-1K Top-1 accuracy versus compute (GFLOPs per image at 224×224), compared with representative CNN, Transformer, Linear, and Hybrid backbones. (b) Compute scaling of SoLA-Vision with input resolution from $16^2$ to $1280^2$, compared against Swin (linear-complexity windowed attention), DeiT (quadratic global attention), and a SoLA-FullSoftmax variant. Patch size is 4.}
    \label{fig:1}
\end{figure}

\begin{figure}
    \centering
    \includegraphics[width=1\linewidth]{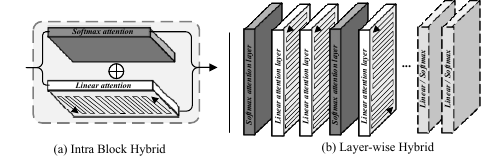}
    \caption{Illustration of (a) Intra-Block Hybrid and (b) Layer-wise Hybrid attention.}    \label{fig:2}
\end{figure}

Given the global modeling capacity of softmax attention and the computational efficiency of linear attention, hybrid architectures have emerged as a natural and promising compromise. However, pioneering efforts \cite{haruna2025vgamba,DBLP:conf/cvpr/HatamizadehK25,han2024agent} typically adopt \textit{rigid intra-block} hybridization strategies(see \Cref{fig:2}). Since each block still executes at least one softmax attention layer, these hybrid blocks are impractical for long-sequence high-resolution inputs and are thus typically deployed only after significant downsampling by a ResNet-style~\cite{he2016deep} CNN stem. Consequently, the potential for more \textit{fine-grained, layer-wise} hybridization within \textbf{attention-only backbones} remains largely unexplored. 

To ground this intuition, we conduct a preliminary experiment using a six-layer stack of standard linear and softmax attention layers. The results shown in \Cref{tab:1.1}, reveal that naively stacking global softmax attention layers leads to performance saturation with a significant increase in computational cost, while inserting just one global softmax layer at a specific layer-wise position into a linear backbone yields considerable performance gains. This raises a critical question: \textit{what is the optimal budget and placement for softmax layers to achieve the best trade-off between performance and computational cost?}

In this work, we analyze the properties of softmax and linear attention layers from the perspective of generalized attention formulation~\cite{choromanski2020rethinking}, identifying the information decay dilemma~(see \cref{fig:3.2}) faced by linear attention in long sequences that restricts long-range features interaction. Drawing parallels to receptive field analysis in CNNs~\cite{luo2016understanding}, our analysis further reveals that the effective interaction radius of stacked linear layers grows only sublinearly with depth. This finding motivates our design principle: global softmax layers act as layer-wise 'shortcuts' between stacked linear attention layers to reestablish long-range dependencies. 

Based on this insight, we propose \textbf{SoLA-Vision (Softmax-Linear Attention Vision)}, an \textbf{attention-only} framework derived from our study on the fine-grained, \textbf{layer-wise hybridization} of linear and softmax attention. We conducted systematic experiments to validate the optimal placement and proportion of global softmax layers within stacked linear attention layers, confirmed that sparsely inserting these global softmax attention layers in the middle and later stages achieves the best performance-cost trade-off. Furthermore, to mitigate the quadratic scaling limitations of global softmax attention in high resolution stages, we design the Hidden State Bridge (HSB), a mechanism that connects the hidden states of high resolution linear attention layers to deeper softmax layers for periodic feature refinement. SoLA-Vision demonstrates strong performance across a range of vision tasks. As shown in \Cref{fig:1}(a), for ImageNet-1K classification, SoLA achieves highly competitive Top-1 accuracy, outperforming strong counterparts among previous CNN, Transformer, Linear, and Hybrid backbones at comparable scales, while maintaining linear computational scaling with input resolution (\Cref{fig:1}(b)). This strong performance extends to dense prediction tasks on COCO~\cite{lin2014microsoft} and ADE20K~\cite{zhou2017scene}, SoLA-Vision surpasses the state-of-the-art hybrid attention model~\cite{DBLP:conf/cvpr/HatamizadehK25} and other strong baseline model by a considerable margin. The main contributions are:

\begin{table}[t]
\centering
\small
\begin{tabular}{cllcc}
\toprule
Model & Pattern & FLOPs & Parameters & Top-1\\
\midrule
Full Linear & LLLLLL & 3.1G & 3.3M& 83.3\\
Full Softmax & SSSSSS & 5.3G & 2.9M & 81.5\\
Softmax First & SLLLLL & 3.4G  & 3.2M& 83.7\\
Softmax middle & LLLSLL & 3.4G &  3.2M& 84.0\\
Softmax Last & LLLLLS & 3.4G &  3.2M& 83.9\\
\bottomrule
\end{tabular}

\caption{Layer-wise hybridization for a six-layer model comparing standard global softmax self-attention (\textbf{S})~\cite{dosovitskiy2020image} and linear attention (\textbf{L})~\cite{duan2024vision}, \textbf{with identical FFN modules}. Models are trained for 300 epochs on the ImageNet-1K 100-class subset. FLOPs are measured per image at $512\times512$ input resolution.}

\label{tab:1.1}
\end{table}


\textbf{(1)} We present a analytical study of the properties of linear and softmax attention in vision backbones, and conduct a systematic empirical study on their hybridization. Our combined results validate that the fine-grained layer-wise hybridization strategy is a simple but highly effective approach for balancing model performance and computational cost.

\textbf{(2)} We introduce \textbf{SoLA-Vision}, an optimized attention-only framework featuring fine-grained, layer-wise hybridization. It is augmented by the \textbf{Hidden State Bridge (HSB)}, a novel mechanism that injects high-resolution linear states into deeper softmax layers for periodic feature refinement.

\textbf{(3)} The SoLA-Vision family (SoLA-T, SoLA-S, SoLA-B) achieves highly competitive performance on ImageNet-1K, COCO, and ADE20K, consistently outperforming prominent Linear, Transformer, Hybrid, and CNN-based counterparts under comparable model scales.

\section{Related works}

\textbf{Transformer Models (Softmax Attention).} Vision Transformers (ViTs)~\cite{dosovitskiy2020image} established softmax self-attention as a powerful tool for visual modeling, leveraging its capacity for precise global context aggregation. Hierarchical architectures like PVT~\cite{wang2021pyramid} and CrossFormer~\cite{wang2023crossformer++} further adapted this paradigm for dense prediction tasks. The primary drawback of this mechanism is its \textbf{quadratic computational complexity ($\mathcal{O}(N^2)$)} relative to the number of input tokens $N$, which becomes prohibitive when processing high-resolution images~\cite{10.1145/3530811}. A dominant strategy to mitigate this bottleneck is \textbf{window-based attention}, employed by models such as the Swin Transformer~\cite{liu2021swin} and Longformer~\cite{beltagy2020longformer}. These methods compute self-attention exclusively within non-overlapping local partitions. While computationally efficient, this local-only approach constrains the effective receptive field at each layer and forcing the model's capacity for global dependency modeling to be contingent on its depth~\cite{yang2021focal}.

\textbf{Linear Attention Models.} To mitigate the quadratic computational complexity of the softmax attention mechanism, an early line of work, such as AFT~\cite{zhai2021attention} and Linformer~\cite{wang2020linformer}, explored feature aggregation mechanisms with \textbf{linear computational complexity}. These models process sequences via a \textbf{compact hidden state}, thus avoiding the explicit materialization of the $N^2$ attention matrix. Furthermore, linear architectures build with linear attention mechanism, such as Mamba~\cite{gu2024mamba}, RWKV~\cite{peng2023rwkv}, and Delta-Net~\cite{yang2024gated}, have emerged as compelling alternatives to Transformers. These architectures have recently been adapted for vision tasks. For instance, Vision-RWKV (VRWKV)~\cite{duan2024vision}, Vim~\cite{zhu2024vision} and Vmamba~\cite{liu2024vmamba} extend the linear model to visual data by introducing multi-directional scanning modules to capture spatial context. However, despite their efficiency, these purely linear designs often exhibit a performance gap compared to Transformers, particularly in capturing fine-grained 2D spatial details and long-range dependencies~\cite{10.1145/3530811}. A hypothesized cause for this limitation is the \textbf{information decay}~\cite{sun2023retentive} inherent in their compressed state representation, where contextual information from distal tokens can be progressively attenuated over a long sequence.

\textbf{Hybrid Attention Models.}
To combine the efficiency of linear attention with the global modeling strength of softmax attention, hybrid architectures have emerged as a compelling research direction. In NLP, models such as Jamba~\cite{lieber2024jamba} have shown evidence that \emph{mixing} linear and softmax attention can lift purely linear language models toward Transformer-like capabilities at a fraction of the computational cost. This paradigm is also gaining traction in industrial deployments (e.g., Minimax-01~\cite{li2025minimax}, Kimi-Linear~\cite{team2025kimi}) and is viewed as a promising avenue for next-generation large models. By contrast, the exploration of such hybrid designs in computer vision remains relatively limited. Prior CV work such as \emph{vGamba}~\cite{haruna2025vgamba} and \emph{Agent attention}~\cite{han2024agent} integrates a linear attention path and a global softmax attention path \emph{within} a single block, forming an \emph{intra-block hybrid}. While this plug-and-play design is convenient, each hybrid block still executes at least one global softmax, which means quadratic costs persist across the network. A notable recent effort, \emph{MambaVision}~\cite{DBLP:conf/cvpr/HatamizadehK25}, adopts a \emph{stage-level hybrid} that combines CNNs with linear attention and softmax attention in a four-stage backbone. The first two stages use residual convolutional blocks to aggressively downsample the input and shorten the sequence length, whereas the later stages employ both MambaVision (linear attention) and Transformer(softmax attention) layers with a roughly half–half split within the stack, yielding strong image-classification performance. However, this stage-level scheme is relatively rigid and includes a substantial proportion of softmax layers. Thus, at higher input resolutions, the softmax attention is implemented in a windowed (local) manner (e.g., window sizes 14 and 7 in stages 3 and 4), which constrains the global feature interaction across the feature map.

In contrast, SoLA-Vision is an attention-only framework. Rather than adhering to rigid templates, we shows that strategically and sparsely inserting a small number of global (non-windowed) softmax layers into an otherwise linear backbone is sufficient to mitigate information decay and restore explicit long-range coupling. This fine-grained layer-wise schedule yields a more favorable accuracy–compute trade-off than prior hybrid designs and, delivers strong results across image classification and dense prediction without introducing additional architectural complexity.

\section{Preliminaries}
\label{sec:prelim}

In this section, we review the properties of softmax and linear attention in vision backbones, providing the theoretical foundation for our hybrid strategy. Let $X\!\in\!\mathbb{R}^{N\times D}$ be an input sequence of $N$ tokens with dimension $D$. We derive queries, keys, and values via linear projections:
$Q=XW_Q, K=XW_K, V=XW_V$.
The projection matrices $W_{Q,K,V}\!\in\!\mathbb{R}^{D\times d}$ map the input to a hidden dimension $d$, yielding $Q,K,V\!\in\!\mathbb{R}^{N\times d}$. We adopt a row-vector convention, where $q_t, k_t, v_t \!\in\!\mathbb{R}^{1\times d}$ denote the $t$-th row vectors. All dot-product scaling factors and normalization layers are omitted for notational clarity.

\subsection{Softmax Attention and Its Generalization}
The standard softmax attention, often referred to as scaled dot-product attention, computes the output as a weighted sum of the value vectors:
\begin{equation}
\label{eq:softmax-token}
\mathrm{SoAttn}_t \;=\;
\frac{\sum_{i=1}^{N}\exp\!\big(q_t k_i^\top\big)\,v_i}
     {\sum_{i=1}^{N}\exp\!\big(q_t k_i^\top\big)}.
\end{equation}
The core of this operation is the attention matrix $A=QK^\top\in\mathbb{R}^{N\times N}$, which explicitly computes a similarity score for every pair of tokens. Consequently, for each token output $\mathrm{SoAttn}_t$, global token interactions are \emph{not} constrained by relative distance. We also write the matrix form inline for compactness as $\mathrm{SoAttn}(Q,K,V)=\mathrm{softmax}\!\big(QK^\top\big)\,V$, where the softmax is applied row-wise. However, computing $N^2$ pairs of attention similarities incurs $\mathcal{O}(N^2 d)$ time and $\mathcal{O}(N^2)$ memory complexity.

\medskip\noindent
A more general perspective replaces the dot product with a nonnegative similarity kernel~\cite{choromanski2020rethinking}
\(\kappa:\mathbb{R}^{d}\times\mathbb{R}^{d}\to\mathbb{R}_{\ge0}\).
Writing \(\langle Q,K\rangle\in\mathbb{R}^{N\times N}\) with entries
\([\langle Q,K\rangle]_{ti}=\kappa(q_t,k_i)\), the normalized kernel attention is
\begin{equation}
\label{eq:gen-attn}
\mathrm{Attn}_t(Q,K,V)
\;=\;
\frac{\sum_{i=1}^{N}\kappa(q_t,k_i)\,v_i}
     {\sum_{i=1}^{N}\kappa(q_t,k_i)}.
\end{equation}
This kernelized view provides a unified perspective, in which Softmax attention is a special case of \eqref{eq:gen-attn} with the choice $\kappa(q,k)=\exp(q k^\top)$.

\subsection{Linear Attention and Information Decay}
\label{sec:prelim:linear-decay}
\begin{figure}
    \centering
    \includegraphics[width=1\linewidth]{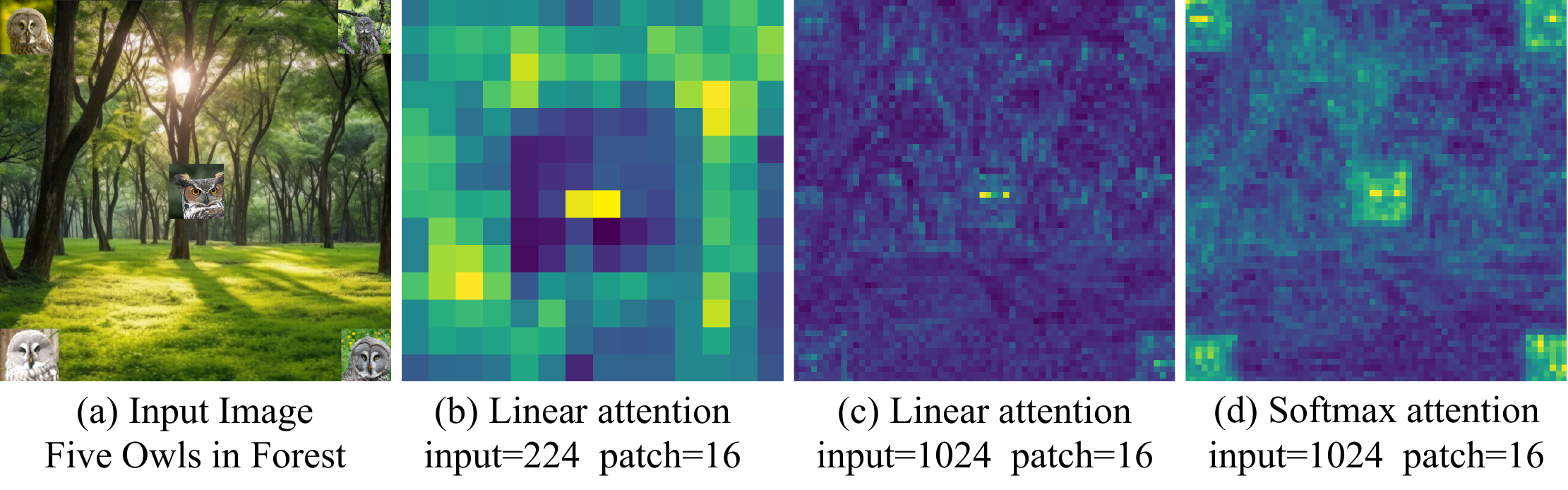}
    \caption{Feature activation maps of pretrained linear and softmax attention layers.}    
    \label{fig:3.2}
\end{figure}
When we choose a \emph{decomposable} kernel $\kappa(q_t,k_i) = \varphi(q_t) \phi(k_i)^\top$ with feature maps $\varphi,\phi:\mathbb{R}^{d}\!\to\!\mathbb{R}^{m}_{\ge0}$, we obtain \emph{linear attention} with $\mathcal{O}(N)$ complexity. The associative property yields $\mathrm{LiAttn}(Q,K,V) =\bigl(\varphi(Q)\phi(K)^\top\bigr)V =\varphi(Q)\bigl(\phi(K)^\top V\bigr)$, which avoids the explicit materialization of the $N^2$ attention matrix. The per-token computation can then be written as:
\begin{align}
\mathrm{LiAttn}_t
&=\sum_{i=1}^{N}\bigl(\varphi(q_t)\phi(k_i)^\top\bigr)\,v_i
 =\varphi(q_t)\!\Bigl(\sum_{i=1}^{N}\phi(k_i)^\top\,v_i\Bigr)\nonumber\\
&=\varphi(q_t) H_t.\label{eq:lin-token}
\end{align}
where $H_t\in\mathbb{R}^{m\times d}$ is the hidden state. 

In modern linear attention architectures, the hidden state $H_t$ is often implemented by a recurrent scan~\cite{yang2024gated} formulated as $\,H_t=\mathrm{decay_t}\odot H_{t-1}+\phi(k_t)^\top v_t$, which runs in linear $\mathcal{O}(Nd)$ computational and memory cost. The update is governed by a key component: the \emph{decay} mechanism that suppresses stale information in the hidden state to mitigate state saturation. Although modern linear-attention architectures make efforts to design selective decay mechanisms, the inherent recurrent form of the update (despite parallel scan implementations) still imposes an implicit \emph{relative-distance dependence} on the decay process. Equivalently, we can write $H_t$ in a parallel (summation) form as distance dependent weighted sum:
\begin{equation}
\label{eq:parallel-Ht}
H_t=\!\!\sum_{\substack{i=1\\ i\neq t}}^{N} \mathrm{decay}\!\big(|t{-}i|\big)\,\phi(k_i)^\top v_i
+\phi(k_t)^\top v_t .
\end{equation}

While reducing the quadratic computational and memory cost, the decay term introduces a critical issue of \emph{information decay}: contributions from distant tokens (at distance $\Delta=|t-i|$) diminish rapidly, attenuating long-range interactions and causing a representational bottleneck. To quantify this attenuation, we adopt the exponential decay model from RetNet~\cite{sun2023retentive}, $\mathrm{decay}(\Delta)=e^{-w\Delta}$, for a scalar decay rate $w>0$. We then define the \emph{effective feature interaction range} $\xi$ as the distance at which this decay factor drops to a small tolerance $\varepsilon>0$. Setting $\mathrm{decay}(\xi)=e^{-w\xi}=\varepsilon$ allows us to solve for $\xi$:
\begin{equation}
\label{eq:effective_range}
\xi\approx\frac{\ln(1/\varepsilon)}{w}.
\end{equation}

In effect, linear attention can be viewed as a soft-boundary windowed aggregation: global visibility is retained, but interactions are mediated by a distance-decay kernel, producing an effective feature interaction range of $\xi$. We empirically visualize this behavior in \Cref{fig:3.2}. At a short sequence length ($N{=}196$, $224\times224$), the linear block produces activations at all five owl locations (\Cref{fig:3.2}(b)). As the sequence scales to $N{=}4096$ ($1024\times1024$), its responses become more localized and some instances are attenuated or vanish (\Cref{fig:3.2}(c)), consistent with distance-dependent decay and a bounded effective receptive field. By contrast, the softmax block at the same long sequence length maintains strong, spatially distributed activations covering all instances (\Cref{fig:3.2}(d)), indicative of stronger global interaction.

\begin{figure*}[t]
    \centering
    \includegraphics[width=1\linewidth]{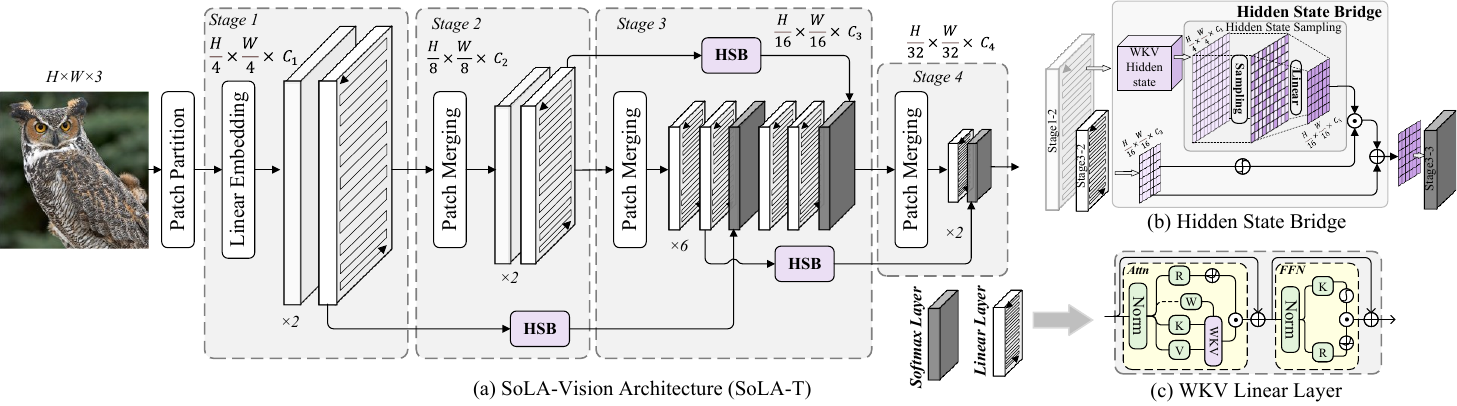}
    \caption{Overall architecture of \textbf{SoLA-Vision}. (a) The SoLA-Vision backbone (SoLA-T shown) with \emph{Hidden State Bridge} (HSB) connections indicated. Stages 1–2 use only linear attention, whereas Stages 3–4 combine linear attention with a small budget of global softmax self-attention layer via fine-grained, layer-wise placement. (b) The Hidden State Bridge mechanism. (c) The WKV linear attention layer architecture.}

    \label{fig:SoLA}
\end{figure*}


\section{Method}

\subsection{Layer-wise Attention Hybridization Study}\label{sec:4.1}

\paragraph{Interaction range scaling in stacked linear attention.}
We follow the analyses of previous CNN receptive field study~\cite{luo2016understanding}, which shows that repeated convolutions develop a Gaussian-like central profile whose variance adds with depth; empirically, even at moderate depths (e.g., 5 layers), the central region already appears close to Gaussian. In our work, according to Eq.~\eqref{eq:parallel-Ht}, we idealize a single linear attention layer as a translation invariant operator with an exponential decay kernel~\cite{sun2023retentive}, an abstraction for linear attention modules:
\begin{equation}
\label{eq:4.1}
K(\Delta)\propto e^{-w|\Delta|}, \qquad K^{(M)}={K_1 * \cdots * K_M}.
\end{equation}
Each exponential decay kernel admits a normalized form $\kappa_{w}(\Delta)=\tfrac{w}{2}e^{-w|\Delta|}$. By the (local) central limit theorem for convolution powers of finite-variance kernels, the \emph{central lobe} of $K^{(M)}$ is well approximated by
\begin{equation}
\label{eq:4.3}
K^{(M)}(\Delta)\;\approx\;\exp\!\Big(-\frac{\Delta^2}{2\sigma_M^2}\Big),\qquad
\sigma_M^2 \;=\; \sum_{\ell=1}^M \frac{2}{w_\ell^2}.
\end{equation}
allowing per-layer rates $w_\ell$. In the identical-layer case $w_\ell\equiv w$, this reduces to $\sigma_M^2=2M/w^2$. Defining the effective interaction radius $\xi_L$ as the distance where the magnitude falls below a small tolerance $\varepsilon>0$, one obtains
\begin{equation}
\label{eq:4.4}
\xi_M \;\approx\; \sigma_M\,\sqrt{2\ln(1/\varepsilon)} \;=\; \mathcal{O}\!\Big(\tfrac{\sqrt{M}}{w}\Big).
\end{equation}
Thus, the effective feature interaction range of stacked linear attention (central lobe) grows only as $\mathcal{O}(\sqrt{M})$, implying diminishing marginal gains from depth alone. This finding directly motivates our design: inserting a few global softmax layers serves as a "shortcut" that reintroduces explicit all pairs coupling after downsampling.


\begin{figure}
    \centering
    
    \includegraphics[width=0.85\linewidth]{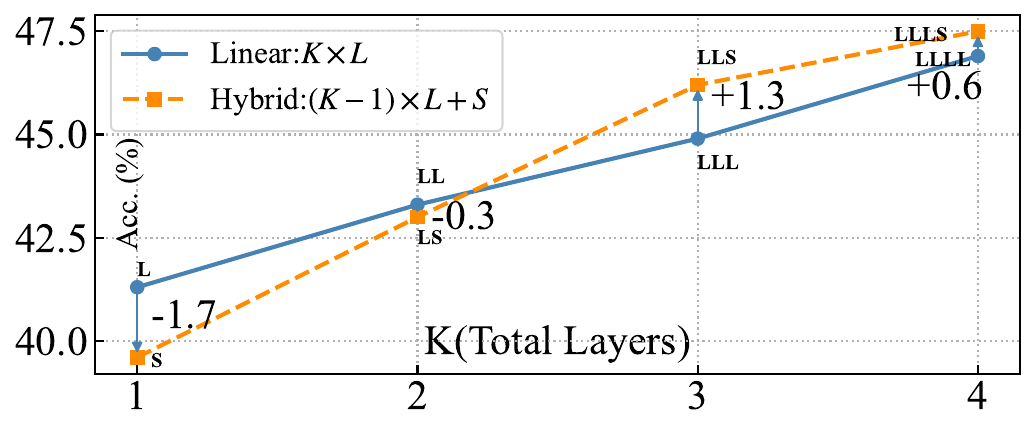}
    \caption{\textbf{Linear and Hybrid Stacking Performance} $K$ denotes the total layers of model. Layers are same as \Cref{tab:1.1} and are \textbf{trained for 300 epochs on the ImageNet-1K 100-class subset.} }
    \label{fig:ratio}
\end{figure}

\paragraph{Empirical study of hybridization schedules.}

Our experiment in \Cref{tab:1.1} (see Intro) shows that inserting softmax attention layers in the middle and later stages of the model yields greater benefits. To further investigate this mixing ratio, we analyze the effect of stacking depth $K$, comparing a pure linear stack ($K \times $\textbf{L}) against a hybrid with a single global softmax layer ($(K-1)\times$ \textbf{L} + \textbf{S}), results are visualized in \Cref{fig:ratio}. Combining with~\Cref{tab:1.1}, we observe that the pure linear attention model consistently outperforms the pure softmax attention model (e.g., S$<$L; SSSSSS$<$LLLLLL) under our experiment setting. This can be attributed to the strong \emph{local inductive bias} conferred by linear attention's distance decay (Sec.~\ref{sec:prelim:linear-decay}), which acts as a spatial prior for vision tasks, \textbf{analogous to} convolutions in a CNN. \textbf{In contrast,} standard softmax attention lacks this bias and must learn all spatial relations from scratch. This insight also explains why hybrid attention does not necessarily produce superior performance (e.g., LS $<$ LL). Only after the stacked linear layers have established local feature representations (leveraging their inductive bias) does inserting a global softmax attention layer yield a significant benefit (e.g., LLS $>$ LLL by 1.3\%).

\subsection{Overall Architecture of SoLA-Vision}

In this part, we introduce SoLA-Vision (SoLA-T, SoLA-S, SoLA-B), \textbf{a family of} attention-only hierarchical backbones with a fine-grained, layer-wise hybrid of linear and softmax attention. Our principle is to preserve the advantages of linear attention and strategically insert \textbf{minimal} global softmax self-attention layers to achieve a trade-off between accuracy and efficiency. SoLA adopts a four-stage hierarchical design like Swin~\cite{liu2021swin}, where each stage contains several attention layers \textbf{scheduled according to our hybrid pattern analysis}. An overview of the SoLA-T architecture is illustrated in \Cref{fig:SoLA}a, detailed configurations are provided in the Appendix.

Specifically, given an input image $I \in \mathbb{R}^{H\times W\times 3}$, it is first partitioned into non-overlapping patches by a stem module, resulting in a 2D feature map with a spatial dimension of $\frac{H}{4} \times \frac{W}{4}$. After incorporating absolute positional embeddings, these tokens are processed through a series of network stages building a feature hierarchy at resolutions of $\frac{H}{8} \times \frac{W}{8}$, $\frac{H}{16} \times \frac{W}{16}$, and $\frac{H}{32} \times \frac{W}{32}$. Each stage comprises a patch merging layer for downsampling (except for the first stage), followed by a stack of attention layers configured according to our layer-wise hybrid pattern. According to our study, placing softmax layers in early high-resolution stages brings limited gains but large computational and memory cost. Thus, Stages 1 and 2 are composed entirely of efficient linear-attention layers. In Stages 3 and 4, we interleave a minimal number of global (non-windowed) softmax layers among linear layers to restore explicit global pairwise interactions after the sequence length has been condensed. \textbf{To compensate for the absence of global softmax in the high-resolution early stages,} we apply the \textbf{Hidden State Bridge (HSB)}. This mechanism injects cross-scale features from the high-resolution linear attention hidden states into deeper global softmax layers, \textbf{thereby} amplifying early-stage global feature interaction and capturing cross-scale dependencies.

\begin{table}[t]
\centering
\caption{Layer-wise hybridization schedules for \textbf{SoLA} Stage~3 (six layers). $k$ denotes the number of global softmax ($S$) layers versus linear ($L$) layers. For this ablation study, stage1,2,4 contain only linear layer. \textbf{Trained for 200 epoch on ImageNet-1K 100-class subset.}}\label{tab:4.1}
\small
\setlength{\tabcolsep}{4pt}
\begin{tabular}{c l c | c l c}
\toprule
$k$ & \textbf{Pattern} & \textbf{Top-1 (\%)} & $k$ & \textbf{Pattern} & \textbf{Top-1 (\%)} \\
\midrule
0 & LLLLLL & 87.53 & 6 & SSSSSS & 88.02\\
\midrule
1 & SLLLLL & 87.62 & 1 & LLLLLS & 87.77 \\
2 & SSLLLL & 88.06 & 2 & LLLLSS & 88.18 \\
3 & SSSLLL & 87.92 & 3 & LLLSSS & 88.02 \\
\midrule
2 & LLSLLS & 88.53 & 3 & LSLSLS & 88.26\\
\bottomrule
\end{tabular}
\end{table}

\subsection{Hybrid Pattern for SoLA}

Building on layer-wise attention hybridization principles in~\Cref{sec:4.1}, we conduct a systematic ablation to identify the optimal concrete implementation within our SoLA backbone. To isolate the effect, this study varies only the six layers of Stage 3; Stages 1, 2, and 4 remain composed entirely of linear attention layers. The results in \Cref{tab:4.1} confirm the superiority of a sparse mixture: the interleaved \texttt{LLSLLS} pattern (88.53\%) significantly outperforms both the pure linear (\texttt{LLLLLL}, 87.53\%) and pure softmax (\texttt{SSSSSS}, 88.02\%) baselines for this stage. Furthermore, placement is critical: for the same budget of $k{=}2$, the interleaved \texttt{LLSLLS} (88.53\%) markedly outperforms a consecutively-stacked \texttt{SSLLLL} (88.06\%). Finally, the budget saturates quickly; increasing to $k{=}3$ (\texttt{LSLSLS}, 88.26\%) shows diminishing returns, confirming that a minimal budget of $k{=}2$ provides the optimal cost-performance trade-off. These findings robustly motivate our default SoLA design: we adopt the \texttt{LLSLLS} schedule in Stage 3.

\subsection{Layer Architecture}

\paragraph{Linear Attention Layer}
For the linear attention layer, we adopt the bi-directional \textit{WKV} attention~\cite{duan2024vision}, a mechanism derived from the RWKV architecture~\cite{peng2023rwkv} (see \Cref{fig:SoLA}c). WKV originates from AFT~\cite{zhai2021attention} and introduces channel-wise decay parameters ($w, u$) to govern the attention computation. As derived in Sec.~\ref{sec:prelim:linear-decay}, this is a specific instantiation of the generalized linear-attention state (Eq.~\eqref{eq:parallel-Ht}) using an exponential kernel ($\phi(x)=\exp(x)$) and a distance-based decay. Its normalized formulation is:
\begin{equation}
\label{eq:wkv}
wkv_t = \frac{ \sum_{i=1,i\neq t}^{N} e^{-(|t-i|-1)/N \cdot w + k_i} v_i + e^{u+k_t} v_t }{ \sum_{i=1,i\neq t}^{N} e^{-(|t-i|-1)/N \cdot w + k_i} + e^{u+k_t} }
\end{equation}
where $N=HW/p^{2}$ is the number of tokens.

The WKV linear attention layer (\Cref{fig:SoLA}c) comprises two sequential sub-modules: a Spatial Mix module and a Channel Mix module. The input $X \in \mathbb{R}^{N \times D}$ first passes through Layer Normalization, and is then projected into receptance ($R_s$), key ($K_s$), and value ($V_s$). The WKV attention (\Cref{eq:wkv}) computes the global context, which is subsequently modulated by the sigmoid-gated receptance $\sigma(R_s)$. The Channel Mix module follows, performing feature fusion via an efficient gated MLP. It also employs Layer Normalization, followed by linear projections to $R_c$ and $K_c$. The value $V_c$ is obtained from $K_c$ using a squared ReLU activation ($\mathrm{ReLU}^2(K_c)$), and the final module output is formed by $\sigma(R_c) \odot V_c$.

\paragraph{Softmax Attention Layer}
For the softmax attention layer, we adopt the standard multi-head self-attention (MHSA) mechanism from the Vision Transformer (ViT)~\cite{dosovitskiy2020image}, where the subsequent MLP block is augmented with a $3\times3$ depth-wise convolution.

\subsection{Hidden State Bridge}
Applying full softmax attention in high-resolution shallow stages is computationally prohibitive and yields diminishing returns. We thus introduce the \textbf{Hidden State Bridge (HSB)} (\Cref{fig:SoLA}b) to provide global refinement for these features at minimal cost.
\begin{equation} \label{eq:hsb}
\begin{aligned}
X_{\text{HSB}} &= \mathrm{Samp}\!\big(H_{(1,2)}, \xi_{\mathrm{eff}}\big)\, W_{\mathrm{HSB}}
\;\in\; \mathbb{R}^{T_3 \times C_3},\\
X_{\text{in}(3,2)} &= X_{\text{out}(3,1)} \;+\; \sigma\!\big(W_g X_{\text{out}(3,1)}\big)\odot X_{\text{HSB}} .
\end{aligned}
\end{equation}
As shown in~\Cref{eq:hsb}, HSB operates by sampling tokens from a high-resolution linear hidden state (e.g., $H_{(1,2)}$) and injecting them into a deeper softmax layer (e.g., $X_{\text{in}(3,2)}$). The design of our sampling function, $\mathrm{Samp}$, is directly motivated by the properties of the linear hidden state. As analyzed in \Cref{sec:prelim:linear-decay}, the hidden state $H$ of a linear attention layer already exhibits local aggregation. Due to the inherent distance-dependent decay, each token effectively encapsulates and represents features from its immediate neighborhood. This property makes sparse sampling a highly effective and efficient strategy. We therefore employ a lightweight, parameter-free equidistant sampling method to select tokens for the bridge. These sparsely sampled tokens are projected by $W_{\mathrm{HSB}}$ to form $X_{\mathrm{HSB}}$ (matching the deeper stage's dimensionality, e.g., $\mathbb{R}^{T_3 \times C_3}$). This bridged representation is then fused with the deep layer's input features (e.g., $X_{\text{out}(3,1)}$) via element-wise addition, modulated by an adaptive gate $\sigma\!\big(W_g X_{\text{out}(3,1)}\big)$ computed directly from the input features.

\section{Experiments}
In this section, we present extensive experiments to evaluate the performance of SoLA-Vision. We compare our models against popular benchmarks covering representative backbone families, from convolutional networks (CNNs) to various attention-based architectures (global softmax, windowed, linear, and hybrids). The model's effectiveness is validated on three standard tasks: imagenet classification, object detection, and semantic segmentation. Params (M) and GFLOPs (G) are measured for the \emph{backbone only}. All experiments were conducted on a server with 8$\times$ NVIDIA RTX-3090 GPUs.

\begin{table}
\centering
\caption{\textbf{ImageNet-1K} image classification comparison with \textbf{representative backbones}. Input resolution is $224\times224$.}

\label{tab:imagenet1k_comparison}
\small
\setlength{\tabcolsep}{4.5pt}
\resizebox{\linewidth}{!}{
\begin{tabular}{lcccc}
\toprule
Method & Type & GFLOPs  & Params(M)  & Top-1(\%)  \\

\midrule
ResNet-18~\cite{he2016deep} & Conv & 1.8   & 11.7   & 69.9 \\
DeiT-T~\cite{touvron2021training}    & Softmax & 1.3   & 5.7    & 72.2 \\
PVT-T~\cite{wang2021pyramid}     & Softmax & 1.9   & 13.2   & 75.1 \\
EffFormer-L1~\cite{li2022efficientformer} & Softmax & 1.31 & 12.3  & 79.2 \\
Vim-T~\cite{zhu2024vision}           & Linear & --   & 7.0    & 76.1 \\
VRWKV-T~\cite{duan2024vision}   & Linear & 1.2   & 6.2    & 75.1 \\
EffVMamba-S~\cite{pei2025efficientvmamba}     & Linear & 1.3   & 11    & 78.7 \\
\rowcolor{solaHL}
\textbf{SoLA-T(ours)} & Hybrid & 1.89  & 6.59 & \textbf{79.8} \\
\midrule
ConvNeXt-T~\cite{liu2022convnet}        & Conv & 4.5   & 28.6  & 82.0 \\
DeiT-S~\cite{touvron2021training}          & Softmax & 4.6   & 22    & 79.8 \\
PVT-M~\cite{wang2021pyramid}     & Softmax & 6.7   & 44.2   & 81.2 \\
Swin-T~\cite{liu2021swin}          & Softmax & 4.5   & 29    & 81.3 \\
Vim-S~\cite{zhu2024vision}           & Linear & 5.3   & 26    & 80.3 \\
VRWKV~\cite{duan2024vision}   & Linear & 4.6   & 23.8   & 80.1 \\
VMamba-T~\cite{liu2024vmamba}        & Linear & 4.9   & 30    & 82.6 \\
Agent-Swin-T~\cite{han2024agent}    & Hybrid & 4.5 & 29  & 82.6 \\
VGamba-L~\cite{haruna2025vgamba}   & Hybrid & 6.3   & 31.9  & 82.8 \\

MambaVision-T~\cite{DBLP:conf/cvpr/HatamizadehK25}   & Hybrid & 4.4   & 31.8  & 82.3 \\

\rowcolor{solaHL}
\textbf{SoLA-S(ours)} & Hybrid & 5.43  & 30.69 & \textbf{82.9} \\
\midrule
ConvNeXt-B~\cite{liu2022convnet}        & Conv & 15.4  & 88.6  & 83.8 \\
DeiT-B~\cite{touvron2021training}          & Softmax & 17.5  & 86    & 81.8 \\
Swin-B~\cite{liu2021swin}   & Softmax & 15.1  & 87.8   & 83.4 \\
Vim-B~\cite{zhu2024vision}           & Linear & --    & 98    & 81.9 \\
VRWKV-B~\cite{duan2024vision}   & Linear & 18.2  & 93.7   & 82.0 \\
VMamba-B~\cite{liu2024vmamba}    & Linear & 15.4 & 89.0  & 83.9 \\
Agent-Swin-B~\cite{han2024agent}    & Hybrid & 15.4 & 88  & 84.0 \\
MambaVision-B~\cite{DBLP:conf/cvpr/HatamizadehK25}    & Hybrid & 15.0 & 97.7  & \textbf{84.2} \\

\rowcolor{solaHL}
\textbf{SoLA-B(ours)} & Hybrid & 14.96  & 88.26 & 84.1 \\
\bottomrule
\end{tabular}
}
\end{table}

\subsection{Image Classification}

For SoLA-Tiny/Small/Base models, we conduct supervised training from scratch on ImageNet-1K. Following the training strategy and data augmentation of Swin~\cite{liu2021swin}, we use a batch size of 1024, AdamW~\cite{loshchilov2017decoupled} with a base learning rate of 1e-3, and a cosine annealing schedule~\cite{loshchilov2016sgdr} to 1e-5 over 300 epochs.

We report the ImageNet-1K performance of SoLA-T/S/B in \Cref{tab:imagenet1k_comparison}, comparing them against representative CNN, Transformers (softmax attention), purely linear-attention models, and other hybrids. At comparable model scales, our SoLA models demonstrate a strong accuracy-compute trade-off. Specifically, \textbf{SoLA-T} achieves \textbf{79.8\%} Top-1 accuracy, substantially outperforming strong linear (e.g., EffVMamba-S, 78.7\%) and softmax (e.g., EffFormer-L1, 79.2\%) counterparts. This trend scales consistently, with \textbf{SoLA-S} achieving \textbf{82.9\%}, surpassing strong baselines like Swin-T (81.3\%) as well as top-performing linear and hybrid competitors like VMamba-T (82.6\%) and MambaVision-T (82.3\%). Finally, \textbf{SoLA-B} achieves a highly competitive \textbf{84.1\%} accuracy. This performance is on par with the state-of-the-art hybrid MambaVision-B (84.2\%) while being significantly more parameter-efficient (88.26M vs. 97.7M for MambaVision-B). These results across all scales validate that our fine-grained, layer-wise hybrid approach effectively balances global context modeling and computational efficiency, achieving a superior performance-cost trade-off compared to prior methods.

\subsection{Downstream Tasks}

\begin{table}
\centering
\caption{Object detection results on \textbf{COCO 2017} using \textbf{Mask R-CNN} (1$\times$ schedule). Input resolution is $1333\times800$. }
\label{tab:coco2017_comparison}
\small

\begin{tabularx}{\columnwidth}{lCCCC} 
\toprule
Method & FLOPs & Params & AP$^{\text{b}}$ & AP$^{\text{m}}$ \\
\midrule
ViT-T~\cite{dosovitskiy2020image} & 147.1G & 8.0M & 41.6 & 37.9 \\
VRWKV-T~\cite{duan2024vision} & 67.9G & 8.4M & 41.7 & 38.0 \\
\rowcolor{solaHL}
\textbf{SoLA-T (ours)} & 56.0G & 6.59M & \textbf{43.8} & \textbf{40.0} \\
\midrule
ViT-S~\cite{dosovitskiy2020image} & 344.5G & 27.5M & 44.9 & 40.1 \\
VRWKV-S~\cite{duan2024vision} & 189.9G & 29.3M & 44.8 & 40.2 \\
\rowcolor{solaHL}
\textbf{SoLA-S (ours)} & 134.0G & 30.69M & \textbf{46.6} & \textbf{42.0} \\
\midrule
ViT-B~\cite{dosovitskiy2020image} & 893.3G & 99.5M & 46.8 & 41.8 \\
VRWKV-B~\cite{duan2024vision} & 599.0G & 106.6M & 46.8 & 41.7 \\
\rowcolor{solaHL}
\textbf{SoLA-B (ours)} & 403.0G & 88.26M & \textbf{47.5} & \textbf{42.3} \\
\bottomrule
\end{tabularx}

\end{table}

\paragraph{Object Detection.}
We evaluate SoLA on the COCO 2017 dataset \cite{lin2014microsoft} using Mask R-CNN \cite{he2017mask} with the 1$\times$ schedule. All backbones are initialized from ImageNet-1K pretraining and trained with an AdamW optimizer (LR 1e-4, weight decay 0.05) and a batch size of 16. As summarized in Table~\ref{tab:coco2017_comparison}, the SoLA models achieve higher box (AP$^{\text{b}}$) and mask (AP$^{\text{m}}$) accuracy than both the global-softmax (ViT) and linear-attention (VRWKV) counterparts, establishing a clear advantage in the accuracy-compute trade-off. This computational gap is primarily because those non-hierarchical baselines require an expensive adapter~\cite{chen2022vision} to generate the multi-scale features that SoLA's native hierarchical design already provides. \textbf{For instance,} the smallest model, \textbf{SoLA-T} (43.8 AP$^{\text{b}}$) achieves a \textbf{+2.2 AP} gain over ViT-T (41.6 AP$^{\text{b}}$) while using \textbf{62\% fewer GFLOPs}. The efficiency gains scale consistently: \textbf{SoLA-B} (47.5 AP$^{\text{b}}$) also surpasses ViT-B (46.8 AP$^{\text{b}}$) in accuracy, while requiring \textbf{55\% fewer GFLOPs}. These results indicate that our sparse, layer-wise hybrid strategy is highly effective and computationally efficient for high-resolution object detection.

\begin{table}
\centering
\caption{Semantic segmentation results on \textbf{ADE20K} using \textbf{UPernet} trained for 160000 iteration. Input resolution is $512\times512$.}
\label{tab:ADE20K}
\small
\setlength{\tabcolsep}{5.75pt}
\begin{tabular}{lcccc}
\toprule
Backbone & Type & FLOPs  & Params  & mIoU  \\
\midrule
ViT-T~\cite{dosovitskiy2020image} & Softmax & 20.9G   & 8.0M   & 42.6 \\
VRWKV-T~\cite{duan2024vision}   & Linear & 16.6G   & 8.4M    & 43.3 \\
EffVMamba-T~\cite{pei2025efficientvmamba}   & Linear & --   & 6.0M    & 38.9 \\
\rowcolor{solaHL}
\textbf{SoLA-T(ours)} & Hybrid & 10.88G  & 6.59M & \textbf{44.7} \\
\midrule
ViT-S~\cite{dosovitskiy2020image} & Softmax & 54.0G   & 27.5M   & 46.2 \\
VRWKV-S~\cite{duan2024vision}   & Linear & 46.3G   & 29.3M    & 47.2 \\
Swin-T~\cite{liu2021swin} & Window & 25.6G   & 28.2M   & 44.5 \\
MambaVision-S~\cite{DBLP:conf/cvpr/HatamizadehK25} & Hybrid & --   & 31.8M   & 46.0 \\
\rowcolor{solaHL}
\textbf{SoLA-S(ours)} & Hybrid & 27.22G  & 30.69M & \textbf{48.1} \\
\midrule
ViT-B~\cite{dosovitskiy2020image} & Softmax & 157.9G   & 99.5M   & 48.8 \\
VRWKV-B~\cite{duan2024vision}   & Linear & 146.0G   & 106.6M    & 49.2 \\
Swin-B~\cite{liu2021swin}   & Window & 87.0G  & 87.8M   & 48.1 \\
MambaVision-B~\cite{DBLP:conf/cvpr/HatamizadehK25} & Hybrid & --   & 97.7M   & 49.1 \\

\rowcolor{solaHL}
\textbf{SoLA-B(ours)} & Hybrid & 89.04G  & 88.26M & \textbf{50.5} \\
\bottomrule
\end{tabular}
\end{table}

\paragraph{Semantic Segmentation.}
We evaluate SoLA on semantic segmentation using the ADE20K dataset~\cite{zhou2017scene} with UPerNet~\cite{xiao2018unified} as the segmentation head. The backbones are initialized from ImageNet-1K pretrained weights and trained with AdamW (initial LR 6e-5, weight decay 0.01), batch size 16, and crop size $512\times512$, for 160k iterations.

As summarized in \Cref{tab:ADE20K}, the SoLA family demonstrates exceptional performance on this dense prediction task. Our strongest comparison comes at the large scale: \textbf{SoLA-B} achieves \textbf{50.5 mIoU}, decisively outperforming the \textbf{Swin-B} baseline (48.1 mIoU) by +2.4 mIoU under a comparable computational and parameter budget. Furthermore, SoLA-B's 89.04G GFLOPs represent a marked reduction compared to ViT-B (157.9G) and VRWKV-B (146.0G). Critically, SoLA also achieves notably higher performance than MambaVision-B (49.1 mIoU), another contemporary hybrid model. The two approaches diverge in their design: MambaVision reverts to \emph{windowed} attention at high resolutions to manage cost~\cite{DBLP:conf/cvpr/HatamizadehK25}, whereas SoLA strategically retains a small number of \emph{global} (non-windowed) softmax layers. This performance gap is consistent with our core design principle, lending strong support to the argument that the retention of explicit, long-range global modeling is particularly beneficial for dense prediction tasks and is the key contributor to SoLA's superior performance.


\subsection{Qualitative Visualization}

\begin{figure}
    \centering
    \includegraphics[width=1\linewidth]{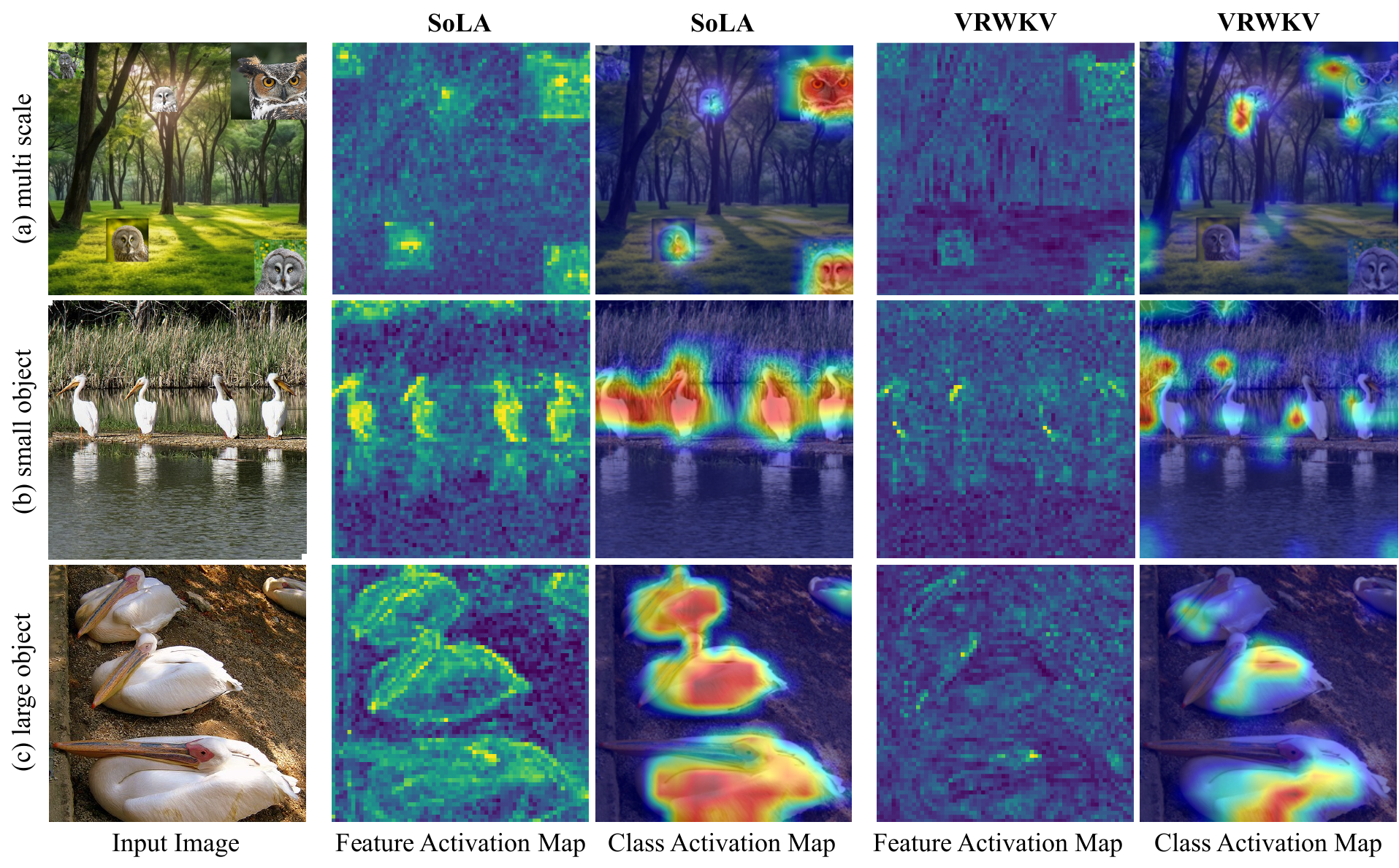}
    \caption{Visualization result. (a) A synthetic image designed to test multi-scale perception. (b, c) Challenging samples selected from the ImageNet-1K validation set. All inputs are resized to $1024 \times 1024$ for analysis.} 
    \label{fig:visualization}
\end{figure}


To better understand how SoLA-Vision processes visual information, we performed a qualitative comparison using visualization methods. We analyze class activation maps using \textbf{EigenCAM} \cite{muhammad2020eigen} and inspect the linear layer feature activation maps to understand the model's internal representations. We compare \textbf{SoLA-T} against the purely linear \textbf{VRWKV-T}, both pretrained on ImageNet-1K. As shown in Fig.~\ref{fig:visualization}, we selected representative images of scenes with small objects, large objects, and multi-scale objects to evaluate the model's perception. The visualized attention of SoLA-T highlights both local features and broader context, successfully localizing all five owls of \textbf{varying scales} (Fig.~\ref{fig:visualization}a) and capturing the full extent of the large cormorant that occupies nearly half the image (Fig.~\ref{fig:visualization}c), whereas the feature maps from the purely linear model (VRWKV-T) show a weaker response of incomplete target activation and missing smaller targets entirely. This demonstrates SoLA-T's superior ability to capture both fine-grained details and long-range dependencies in complex natural scenes, which validates the effectiveness of our flexible layer-wise hybrid attention.

\begin{table}[t]
\centering
\small
\caption{Ablation study of the backbone schedule and the Hidden State Bridge (HSB).}
\label{tab:ablation}
\setlength{\tabcolsep}{5pt} 
\begin{tabular}{l l c}
\toprule
\textbf{Pattern} & \textbf{Layer-wise Schedule} & \textbf{Acc} \\
   & (stage1/stage2/stage3/stage4) &  (\%) \\

\midrule
Alternating & $LL / LL / LSLSLS / LS$ & 79.12 \\
Stacked-Last & $LL / LL / LLLSSS / LS$ & 79.28 \\
\textbf{SoLA Pattern} & \textbf{$LL / LL / LLSLLS / LS$} & 79.68 \\
\textbf{SoLA Pattern + HSB} & \textbf{$LL / LL / LLSLLS / LS$} & 79.83 \\
\bottomrule
\end{tabular}
\end{table}

\subsection{Ablation Study}
We conduct a final ablation study on the complete backbone schedule and the Hidden State Bridge (HSB), with results summarized in \Cref{tab:ablation}. We first validate our proposed \textbf{SoLA Pattern} by comparing it against the "Alternating" and "Stacked-Last" schedules. SoLA with our sparse pattern achieves \textbf{79.68\%} Top-1, while using fewer softmax attention layers. This confirms our design choice. Finally, we integrate the \textbf{HSB}, which further lifts the performance to \textbf{79.83\%} (+0.15\%), validating its effectiveness as a key component of our final architecture.

\section{Conclusion}
\label{sec:conclusion}

Our work presents \textbf{SoLA-Vision}, a fully attention-based backbone that leverages fine-grained, layer-wise hybridization of linear and softmax attention. Informed by systematic analysis of linear and softmax attention hybridization properties, our architecture employs simple yet effective hybrid patterns, achieving stronger accuracy-efficiency trade-offs compared to pure linear, windowed, and previous hybrid backbones. For future work, experiments indicate that the optimal hybrid configuration may depend on the task and model scale. Manually searching for the best placement is inefficient and computationally expensive. Thus, a promising research direction is to develop a \textbf{learnable} policy that can automatically determine the layer-wise assignment of linear and softmax attention, potentially tailoring the architecture to specific tasks, datasets, or computational constraints.

{
    \small
    \bibliographystyle{ieeenat_fullname}
    \bibliography{main}
}



\clearpage
\setcounter{page}{1}
\maketitlesupplementary

\begin{table*}[!t]
\centering
\small
\setlength{\tabcolsep}{4pt}
\renewcommand{\arraystretch}{1.2}
\begin{tabular}{|l|c|c|c|c|}
\hline
  & input size & SoLA-T & SoLA-S & SoLA-B \\
\hline
stem & $224\times224$
     & patch size $4\times4$, dim 96
     & patch size $4\times4$, dim 96
     & patch size $4\times4$, dim 128 \\
\hline

\multirow{2}{*}{stage 1} & \multirow{2}{*}{$56\times56$}
    & \textbf{WKV Linear layer} $\times 2$
    & \textbf{WKV Linear layer} $\times 2$
    & \textbf{WKV Linear layer} $\times 2$ \\

 & & patch merging $96\rightarrow 128$
   & patch merging $96\rightarrow 192$
   & patch merging $128\rightarrow 256$ \\
\hline

\multirow{2}{*}{stage 2} & \multirow{2}{*}{$28\times28$}
    & \textbf{WKV Linear layer} $\times 2$
    & \textbf{WKV Linear layer} $\times 2$
    & \textbf{WKV Linear layer} $\times 2$ \\
 & & patch merging $128\rightarrow 192$
   & patch merging $192\rightarrow 384$
   & patch merging $256\rightarrow 512$ \\
\hline

\multirow{2}{*}{stage 3} & \multirow{2}{*}{$14\times14$}
    & \textbf{WKV Linear layer} $\times 2$
    & \textbf{WKV Linear layer} $\times 2$
    & \textbf{WKV Linear layer} $\times 3$ \\
& & \textbf{ViT Softmax layer} $\times 1$
   & \textbf{ViT Softmax layer} $\times 1$
   & \textbf{ViT Softmax layer} $\times 1$ \\
  & & \textbf{WKV Linear layer} $\times 2$
    & \textbf{WKV Linear layer} $\times 2$
    & \textbf{WKV Linear layer} $\times 2$ \\
& & \textbf{ViT Softmax layer} $\times 1$
   & \textbf{ViT Softmax layer} $\times 1$
   & \textbf{ViT Softmax layer} $\times 1$ \\
& &  6 $Layers$
   & 6 $Layers$
   & ... 16 $Layers$ \\
& &  \textbf{LLSLLS}
   & \textbf{LLSLLS} 
   & \textbf{LLLSLLSLLSLLSLLS} \\

& & patch merging $192\rightarrow 256$
   & patch merging $384\rightarrow 768$
   & patch merging $512\rightarrow 1024$ \\
\hline

stage 4 & $7\times7$
        & \textbf{WKV Linear layer} $\times 1$
        & \textbf{WKV Linear layer} $\times 1$
        & \textbf{WKV Linear layer} $\times 1$ \\
    & & \textbf{ViT Softmax layer} $\times 1$
        & \textbf{ViT Softmax layer} $\times 1$
        & \textbf{ViT Softmax layer} $\times 1$ \\

\hline

\textbf{HSB configs}  & \textbf{(stage, layer)} 
        & $(1, 2)L \rightarrow (3, 3)S$  
        & $(1, 2)L \rightarrow (3, 3)S$
        & $(1, 2)L \rightarrow (3, 3)S$ \\
    & \textbf{L/S} 
        & $(2, 2)L \rightarrow (3, 6)S$
        & $(2, 2)L \rightarrow (3, 6)S$
        & $(1, 2)L \rightarrow (3, 6)S$ \\ 
    &   & $(3, 2)L \rightarrow (4, 2)S$
        & $(3, 2)L \rightarrow (4, 2)S$
        & $(2, 2)L \rightarrow (3, 9)S$ \\
    &   & 
        & 
        & $\;\; (2, 2)L \rightarrow (3, 12)S$ \\
    &   &  
        &  
        & $\;\; (3, 2)L \rightarrow (3, 15)S$ \\
    &   &  
        &  
        & $(3, 2)L \rightarrow (4, 2)S$ \\
   
\hline

Params (M) &  & 6.59 & 30.69 & 88.26 \\
\hline
FLOPs (G)&   & $1.89\times10^{9}$ & $5.43\times10^{9}$ & $14.96\times10^{9}$ \\
\hline
\end{tabular}
\caption{Detailed Architectures of SoLA-Vision-T/S/B.}
\label{tab:sola_arch}
\end{table*}

\section{Detailed Configurations of SoLA-Vision}
\label{sec:detailed_arch}

In Table~\ref{tab:sola_arch}, we provide the detailed architectural configurations of the SoLA-Vision family (SoLA-T, SoLA-S, and SoLA-B). All variants share a four stage hierarchical backbone initialized with a patch-embedding stem, followed by patch-merging downsampling modules. They differ primarily in channel dimensions and the number of layers per stage. The architecture adopts our proposed layer-wise hybrid strategy, which interleaves linear attention layers with ViT-style global softmax self-attention layers. Additionally, the Hidden State Bridge (HSB) mechanism sparsely routes shallow linear features into selected deep softmax layers, allowing SoLA-Vision to reuse early global context without introducing extra quadratic-cost attention at high resolutions.

\section{Additional Ablation Studies}
\label{sec:add_ablation}

\subsection{HSB on Dense Prediction}
\label{subsec:hsb_downstream}
To ensure a fair comparison with baseline backbones, the results reported in the \textbf{Downstream Tasks} section of the main paper were obtained \textbf{without activating the Hidden State Bridge (HSB)}. While the HSB was originally designed for fine-grained linear-softmax attention hybridization, its mechanism of cross-scale transmission inherently benefits dense prediction tasks by recovering high-frequency spatial details often lost during downsampling. In this section, we \textbf{activate} the HSB to demonstrate the full potential of SoLA-Vision in dense prediction tasks.

As reported in \Cref{tab:ablation_hsb}, activating the HSB yields consistent improvements on ADE20K semantic segmentation across all model scales. Specifically, SoLA-T, SoLA-S, and SoLA-B achieve mIoU gains of \textbf{+0.3}, \textbf{+0.5}, and \textbf{+0.9}, respectively. These results validate that HSB effectively boosts dense prediction performance with negligible computational overhead.

\begin{table}[H]
\centering
\caption{\textbf{Ablation of HSB on ADE20K (Semantic Segmentation).} Models are trained with UPerNet for 160k iterations. The HSB module significantly boosts mIoU with zero parameter overhead.}
\label{tab:ablation_hsb}
\small
\setlength{\tabcolsep}{8pt} 
\begin{tabular}{lcccc}
\toprule
Backbone & \textbf{HSB} & FLOPs & mIoU \\
\midrule
SoLA-T & \ding{55} & 10.9G & 44.7 \\
\textbf{SoLA-T} & \ding{51} & 10.9G & \textbf{45.0} \textcolor{green!60!black}{(+0.3)} \\
\midrule
SoLA-S & \ding{55} & 27.2G & 48.1 \\
\textbf{SoLA-S} & \ding{51} & 27.2G & \textbf{48.6} \textcolor{green!60!black}{(+0.5)} \\
\midrule
SoLA-B & \ding{55} & 89.0G & 50.5 \\
\textbf{SoLA-B} & \ding{51} & 89.0G & \textbf{51.4} \textcolor{green!60!black}{(+0.9)} \\
\bottomrule
\end{tabular}
\end{table}

\subsection{Pure Linear Variant of SoLA-Vision}
\label{subsec:pure_linear}

To explicitly verify that the performance gains stem from our layer-wise attention hybrid strategy rather than the hierarchical design of linear attention backbone, we conducted a full-scale evaluation by training a ``Pure Linear'' variant of SoLA-T on ImageNet-1K for 300 epochs. This experiment follows the \textbf{identical training recipe} used in the main paper's ablation studies to ensure a fair comparison. 

As shown in \Cref{tab:ablation_pattern}, replacing all Softmax attention layers with Linear attention results in a significant performance degradation (\textbf{79.68\%} $\rightarrow$ \textbf{78.56\%}). This \textbf{1.12\%} drop confirms that the hierarchical linear design alone is insufficient; the interleaved Softmax layers are indispensable for capturing the global context and precise retrieval capabilities that pure linear models lack.

\begin{table}[H]
\centering
\small
\caption{\textbf{Ablation with Full Linear variant of SoLA-T.} We compare different attention schedules. `L' denotes Linear Attention and `S' denotes Softmax Attention in Stage 3. The Pure Linear variant significantly lags behind the hybrid patterns.}
\label{tab:ablation_pattern}
\setlength{\tabcolsep}{4pt} 
\begin{tabular}{l l c}
\toprule
\textbf{Pattern} & \textbf{Layer-wise Schedule} & \textbf{Acc} \\
& (stage1/stage2/stage3/stage4) & (\%) \\
\midrule
\textbf{Pure Linear} & $LL / LL / LLLLLL / LL$ & \textbf{78.56} \textcolor{red!60!black}{(-1.1)}\\
Alternating & $LL / LL / LSLSLS / LS$ & 79.12 \\
Stacked-Last & $LL / LL / LLLSSS / LS$ & 79.28 \\
\textbf{SoLA Pattern} & {$LL / LL / LLSLLS / LS$} & \textbf{79.68} \\
\bottomrule
\end{tabular}
\end{table}

\section{Further Discussion on the Interaction Range of Stacked Linear Attention}
\label{app:interaction}

In the main paper, we argued that the effective interaction range of stacked linear attention layers grows sublinearly with depth. We follow the receptive field analysis established in previous CNN studies, which examines signal propagation through repeated convolutions. On a 2D pixel lattice, the sequential application of translation-invariant kernels results in an overall influence profile that asymptotically approaches a Gaussian distribution. Crucially, the variance of this distribution accumulates with network depth, limiting the effective interaction range. 

Our setting differs only in geometry. Transformer-style attention backbones operate on a 1D sequence of encoded tokens $(x_1,\dots,x_N)$, which we view as a 1D lattice indexed by $t\in\mathbb{Z}$. Under the standard idealization used in the main text, a single linear attention layer along this sequence acts as a translation-invariant operator with a distance-dependent kernel:
\begin{equation}
\label{eq:app-kernel}
y_t \;=\; \sum_{\Delta\in\mathbb{Z}} K(\Delta)\,x_{t-\Delta},
\qquad
K(\Delta)\propto e^{-w|\Delta|},
\end{equation}
where $\Delta$ is the offset between query and key positions and $w>0$ is a decay rate. \Cref{eq:app-kernel} is exactly a 1D convolution along the token dimension; thus, the convolutional analysis applies directly to this 1D lattice.

We now make this correspondence explicit for $M$ stacked linear attention layers. For each layer $\ell$, we consider the normalized exponential kernel:
\begin{equation}
\label{eq:app-kappa}
K_{w_\ell}(\Delta)
\;=\;
\frac{w_\ell}{2}e^{-w_\ell|\Delta|},
\qquad
\sum_{\Delta\in\mathbb{Z}} K_{w_\ell}(\Delta)=1.
\end{equation}
This kernel is symmetric around $0$ and decays exponentially with $|\Delta|$. A standard variance derivation yields:
\begin{equation}
\label{eq:app-var}
\mathbb{E}_{K_{w_\ell}}[\Delta]=0,
\qquad
\mathrm{Var}_{K_{w_\ell}}[\Delta] \;=\; \frac{2}{w_\ell^2}.
\end{equation}
Stacking $M$ such layers corresponds to convolving these kernels, yielding an effective kernel
$K^{(M)} = K_{w_1} * \cdots * K_{w_M}$. Interpreting $K^{(M)}$ as a discrete probability kernel over offsets, let $\Delta\sim K^{(M)}$ denote the induced displacement and let $\sigma_M^2$ be its variance. Using the zero-mean property and independence across layers, the variances add:
\begin{equation}
\label{eq:app-SM-var}
\sigma_M^2
\;=\;
\mathrm{Var}(\Delta)
\;=\;
\sum_{\ell=1}^M \frac{2}{w_\ell^2}.
\end{equation}

A local central limit theorem for lattice variables implies that, in the central region (i.e., $|\Delta|=o(\sigma_M)$ as $M$ grows), the stacked kernel $K^{(M)}(\Delta)$ is well approximated by a discrete Gaussian with variance $\sigma_M^2$:
\begin{equation}
\label{eq:app-central}
K^{(M)}(\Delta)
\;\approx\;
\exp\!\Big(-\frac{\Delta^2}{2\sigma_M^2}\Big),
\end{equation}
where we omit the normalization constant as we focus on the spatial decay profile.

To quantify the interaction range of this central lobe, we fix a small tolerance $\varepsilon>0$ and define $\xi_M$ as the distance where the magnitude of $K^{(M)}$ drops below $\varepsilon$:
\begin{equation}
\label{eq:app-threshold}
K^{(M)}(\xi_M)
\;\approx\;
\exp\!\Big(-\frac{\xi_M^2}{2\sigma_M^2}\Big)
\;\approx\;
\varepsilon.
\end{equation}
Solving \eqref{eq:app-threshold} gives:
\begin{equation}
\label{eq:app-radius-general}
\xi_M^2
\;\approx\;
2\sigma_M^2\ln(1/\varepsilon),
\qquad
\xi_M
\;\approx\;
\sigma_M\sqrt{2\ln(1/\varepsilon)}.
\end{equation}
In the identical-layer case where $w_\ell\equiv w$, \eqref{eq:app-SM-var} reduces to
\begin{equation}
\label{eq:app-identical-var}
\sigma_M^2 \;=\; \frac{2M}{w^2},
\end{equation}
and substituting this into \eqref{eq:app-radius-general} simplifies to:
\begin{equation}
\label{eq:app-radius-scaling}
\xi_M
\;\approx\;
\sqrt{\frac{4M}{w^2}\ln(1/\varepsilon)}
\;=\;
\mathcal{O}\!\Big(\tfrac{\sqrt{M}}{w}\Big).
\end{equation}
This formalizes the statement in the main paper: under an exponential decay idealization, the effective feature interaction range of stacked linear attention (central lobe) grows only as $\mathcal{O}(\sqrt{M})$ with depth.

\section{Analysis of Feature Activation Maps}
\label{sec:fam_analysis}

We analyze the Feature Activation Maps (FAM) to verify that the distinctive highlighting of salient regions stems from learned semantic representations rather than numerical artifacts introduced by normalization layers. 

\begin{figure}[t]
    \centering
    \includegraphics[width=1\linewidth]{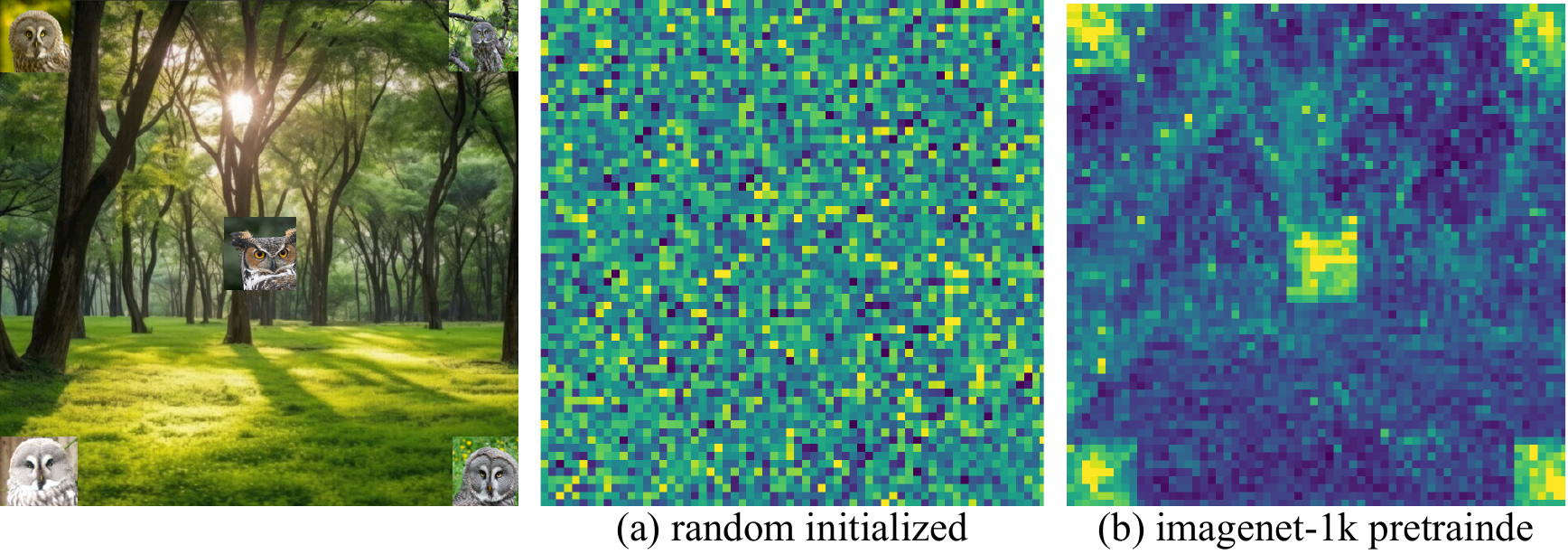}
    \caption{\textbf{Visualization of Feature Activation Maps.} 
    \textbf{(a)} With random initialization, the map shows chaotic noise. This contrast proves that the feature concentration is driven by learned parameters rather than normalization statistics. \textbf{(b)} With ImageNet-1K pre-training, the map highlights semantically relevant regions, demonstrating effective feature learning.}
    \label{fig:fam}
\end{figure}

As depicted in \Cref{fig:fam}, the randomly initialized baseline (\Cref{fig:fam}a) produces disordered, high-entropy patterns despite undergoing the identical normalization operations. In contrast, the model initialized with ImageNet-1K pre-trained weights (\Cref{fig:fam}b) exhibits clear, structurally coherent activations that align precisely with object boundaries. This qualitative discrepancy confirms that the focused activation is a genuine product of the pre-training process, effectively encoding high-level semantic information.

\end{document}